%% file: ACR2165.tex
\documentclass{svproc}
\usepackage{booktabs}
\usepackage{graphicx}
\usepackage[hidelinks]{hyperref} 
\usepackage{cleveref} 
\usepackage{url}

\begin{document}
\mainmatter              
\title{\small{
    This is an Author's Original Manuscript of an article whose final and definitive form, the Version of Record, has been published in the \textit{Proceedings of the Third International Conference on Advances in Computing Research (ACR'25)}, available online at: \href{https://link.springer.com/chapter/10.1007/978-3-031-87647-9_25}{https://link.springer.com/chapter/10.1007/978-3-031-87647-9\_25}
}\\[1em]
 \large{Performance of Machine Learning Classifiers for Anomaly Detection in Cyber Security Applications}}
\titlerunning{Anomaly Detection for Cyber Security Applications}  

\author{Markus Haug\textsuperscript{[0009-0006-9581-6003]*} \and Gissel Velarde\textsuperscript{[0000-0001-5392-9540]}}
\authorrunning{Markus Haug and Gissel Verlade} 

\institute{IU International University of Applied Sciences, Erfurt, 99084, Germany\\
\email{markus.haug@iu-study.org, gissel.velarde@iu.org}}

\maketitle              

\pagenumbering{gobble} 

\begin{abstract}
This work empirically evaluates machine learning models on two imbalanced public datasets (KDDCUP99 and Credit Card Fraud 2013). The method includes data preparation, model training, and evaluation, using an 80/20 (train/test) split. Models tested include eXtreme Gradient Boosting (XGB), Multi Layer Perceptron (MLP), Generative Adversarial Network (GAN), Variational Autoencoder (VAE), and Multiple-Objective Generative Adversarial Active Learning (MO-GAAL), with XGB and MLP further combined with Random-Over-Sampling (ROS) and Self-Paced-Ensemble (SPE). Evaluation involves 5-fold cross-validation and imputation techniques (mean, median, and IterativeImputer) with 10, 20, 30, and 50 \% missing data. Findings show XGB and MLP outperform generative models. IterativeImputer results are comparable to mean and median, but not recommended for large datasets due to increased complexity and execution time. The code used is publicly available on GitHub (\href{https://github.com/markushaug/acr-25}{github.com/markushaug/acr-25}).
\end{abstract}

\input{sections/0_introduction}
\input{sections/2_methodology}

\input{sections/3_results}
\input{sections/4_discussion}
\input{sections/5_conclusion}
\input{sections/6_contributions}
%
%

\bibliographystyle{spmpsci} 
\bibliography{references}

\end{document}

%% file: sections/0_introduction.tex
\section{Introduction}

In this work, we evaluate the performance of several machine learning algorithms for binary classification on tabular data. We compare the performance of supervised and unsupervised learning approaches, and study the case of fraud detection for cyber security. Still, our findings and lessons learned are relevant to any business application and domain where the solution can be framed as a binary classification problem including fraud detection, credit approval, medical diagnosis, or online advertising, to mention a few applications. Indeed, very few business use cases would have balanced data in terms of negative and positive ratios. Generally, the positive class samples appear less frequently than those of the negative class. However, the positive samples are of high interest when diagnosing sicknesses, identifying fraudulent activities or predicting if a customer will click on an ad.  

Binary classification of imbalance data has been a relevant topic for the machine learning community given that it is challenging to learn statistics from the under represented samples \cite{JMLR:v18:16-365}. Approaches to imbalanced learning include hyperparameter tuning \cite{wang2020imbalance,VELARDE2024200354,li2023imbalanced}, sampling the data before training \cite{kim2022,hajek2022fraud,VELARDE2024200354}, and ensembling \cite{yang2021progressive,JMLR:v18:16-365}.

Cyber security applications are becoming increasingly important given that cyber crime is on the rise with cost estimations that reach the 400 billion USD globally~\cite{sarker2020cybersecurity}. For this experimental study we systematically evaluated several algorithms on fraud detection datasets, one of the applications of cyber security, however, the main take aways and findings of this study, apply to any application that can be framed as a binary classification problem. The motivation to compare supervised learning with unsupervised learning algorithms is justified since fraud patterns might change over time, fraudsters are very creative and try new possibilities to bypass any system, and therefore, some patterns may never appear during training or might be recognised in a larger time window when it is too late \cite{VELARDE2024200354}. Therefore, an approach that can discriminate between legit and fraud activity without the need of collecting labels, is very appealing. 

In our experiments, we focus first on the evaluation of different approaches and techniques to deal with imbalanced learning. Here we compare the performance of tree based and deep learning approaches in supervised and unsupervised fashion, and test the effect of sampling and ensembling. Then, we investigate the robustness of the best classifier when dealing with missing data, a problem that occurs either because certain features might not be available at the time of data collection, either because of the nature of the process or due to a system error. In the next section we describe the method. 

%% file: sections/2_methodology.tex
\section{The Method} \label{sec:method}

\begin{figure}[htb!]
    \includegraphics[width=\textwidth]{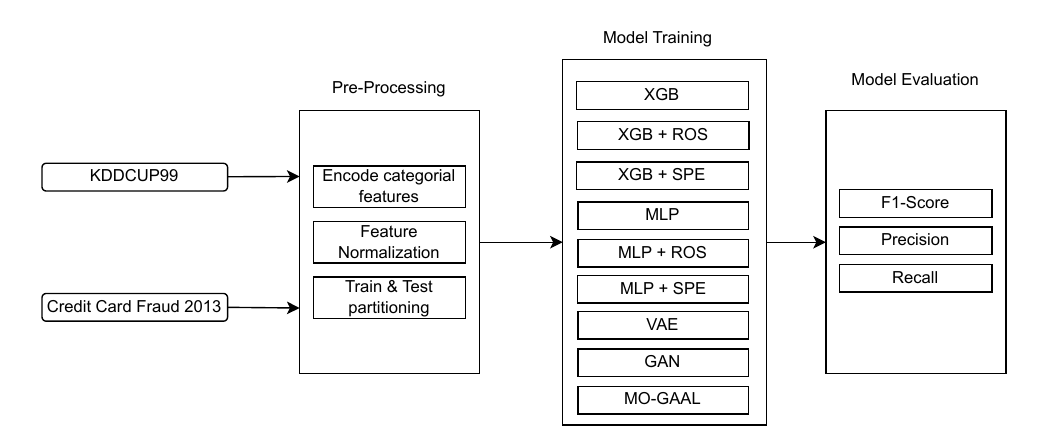}
    \centering
    \caption[The Method]{Graphical representation of the Method. Each dataset goes through a Pre-Processing stage before Model Training and Evaluation. Models are trained and tested independently. XGB and MLP are selected and combined with ROS and SPE.}
    \label{fig:visualized-method}
\end{figure}

The method can be seen in  \Cref{fig:visualized-method}. Each dataset is initially preprocessed in a standardized way so that they can be reused for the various models and sampling methods. The respective properties of the datasets can be seen in \Cref{tab:datasets-characteristics}. Categorical features are first converted into numerical features and then all numerical features are normalized using a \textit{Standard Scaler} \cite{scikit-learn}. 
Model training and evaluation follows a 80/20 (train/test) split. The partitioning is not purely stochastic, but is carried out as part of stratified sampling, so that the respective imbalance ratio (IR) is retained in all created partitions. Models tested include XGB \cite{chenXGBoostScalableTree2016}, MLP \cite{rumelhartLearningRepresentationsBackpropagating1986}, GAN \cite{goodfellowGenerativeAdversarialNets2014}, VAE \cite{bankAutoencoders2021}, and MO-GAAL \cite{winstonliSourceCodePyod2022}, with XGB and MLP further combined with ROS and SPE. The experiment was executed in a 5-fold cross-validation on the training set with scikit-learn pipelines \cite{scikit-learn}. After evaluation, XGB was selected to study the effect of imputation techniques (mean, median, and IterativeImputer) with 10, 20, 30, and 50 \% missing data. This last step is not represented in \Cref{fig:visualized-method}. The evaluation is based on the number of True positive (TP), True negative (TN), False positive (FP), and False negative (FN), considering $Precision$ = $TP/(TP + FP)$, $Recall$ = $TP/(TP + FN)$ and   $F_1$-$Score$ = $2 \times Recall \times Precision/(Recall + Precision)$.  Finally, significance tests are performed between imputation techniques. All experiments were conducted on a 2021 MacBook Pro with the Apple M1 Max and 32 GB of RAM.

\subsection{Datasets} \label{ref:datasets}

We used two highly imbalanced datasets \textit{KDDCUP99} and \textit{Credit Card Fraud 2013}, see Table \ref{tab:datasets-characteristics}. A scikit-learn pipeline was used to intially preprocess the datasets in a standardized way to be reused for the various models and sampling methods. The categorical features were first converted into numerical features and then all numerical features were scaled. Finally, a training and test data partition was created. The partitioning was not purely stochastic, but was carried out as part of stratified sampling, so that the respective IR was retained in all created partitions.
\begin{table}[h]
  \centering
  \small
  \caption{Characteristics of the datasets used. Data from \cite{pozzoloCalibratingProbabilityUndersampling2015,KDDCup19991999}.}
  \resizebox{\textwidth}{!}{%
    \begin{tabular}{p{25mm}p{20mm}p{20mm}p{25mm}p{20mm}p{20mm}p{20mm}}
      \toprule
      \textbf{Dataset} & \textbf{Negative Samples} & \textbf{Positive Samples} & \textbf{IR} & \textbf{NaN Values} & \textbf{Numerical Features} & \textbf{Categorical Features} \\ 
      \midrule
      Credit Card & $284\,315$ & 492 & 577.88 & 0 & 29 & 0 \\ 
      KDDCUP99 & $60\,593$ & $250\,436$ & 4.13 & 0 & 34 & 7 \\
      \bottomrule
    \end{tabular}%
  }
  \label{tab:datasets-characteristics}
\end{table}

\subsection{Hyperparameters}  \label{sec:hyperparams}

 \begin{table}[ht!]
  \centering
  \tiny
  \caption{Hyperparameters found for the Credit Card dataset.}
    \begin{tabular}{p{1.5cm}p{11cm}} \toprule
        \textbf{Model}   & \textbf{Parameters} \\ \midrule
        XGB (Normal) & Vanilla XGB with default parameters from \cite{xgboostdevelopersXGBoostParameters2022} \\ \cmidrule(l){1-2}

          XGB + ROS & objective = Vanilla XGB with default parameters from \cite{xgboostdevelopersXGBoostParameters2022} \\ \cmidrule(l){1-2}

          XGB + SPE & Vanilla XGB with default parameters from \cite{xgboostdevelopersXGBoostParameters2022} \\ \cmidrule(l){1-2}

          MLP (Normal) & batch\_size = 2048, epochs = 50, learning\_rate = 0.01, optimizer = Adam, layers = [256, 256, Dropout(0.3), 256, Dropout(0.3), 1], activation = [ReLU, ReLU, ReLU, sigmoid], metrics = [f1, fn, fp, tn, tp, precision, recall], loss = binary\_crossentropy \\ \cmidrule(l){1-2}

          MLP + ROS & MLP (Normal), sampling\_strategy = 1 \\ \cmidrule(l){1-2}

          MLP + SPE & estimator = MLP (Normal), n\_estimators = 50 \\ \cmidrule(l){1-2}

          VAE & epochs = 10, batch\_size = 32, lr=0.001, dropout\_rate = 0.2 \\ \cmidrule(l){1-2}

          GAN & epochs = 10, latent\_dim = 29, batch\_size = 32, d\_optimizer = SGD with learning rate = 0.0002, g\_optimizer = Adam with learning rate = 0.00001 \& beta\_1 = 0.5 \\ \cmidrule(l){1-2}

          MO-GAAL & contamination = Proportion of fraud cases in training partition, n\_sub\_generators = 3, learning rate discriminator = 0.01, learning rate generator = 0.0001, epochs = 2 \\ \cmidrule{1-2}
        \end{tabular}
          \label{tab:hyperparams-cc}
\end{table}

\begin{table}[ht!]
  \centering
  \tiny
  \caption{Hyperparameters found for the KDDCUP99 dataset.}
    \begin{tabular}{p{1.5cm}p{11cm}} \toprule
        \textbf{Model}   & \textbf{Parameters} \\ \midrule
        XGB (Normal) & objective = binary:logistic, booster = gbtree, colsample\_bytree = 0.7, device = cpu, eval\_metric = aucpr, gamma = 0.3, learning\_rate = 0.3, max\_depth = 6, max\_leaves  = 64, n\_estimators = 1000 \\ \cmidrule(l){1-2}

        XGB + ROS & Vanilla XGB with default parameters from \cite{xgboostdevelopersXGBoostParameters2022}, sampling\_strategy = 1 \\ \cmidrule(l){1-2}

        XGB + SPE & Vanilla XGB with default parameters from \cite{xgboostdevelopersXGBoostParameters2022}, n\_estimators (SPE) = 50 \\ \cmidrule(l){1-2}

        MLP (Normal) & epochs = 50, batch\_size = 2048, learning\_rate = 0.02, optimizer = Adam, layers = [128, 64, 1], activation = [relu, tanh, sigmoid], metrics = [f1, recall, precision], loss = binary\_crossentropy \\ \cmidrule(l){1-2}

        MLP + ROS & estimator = MLP (Normal), sampling\_strategy = 1 \\ \cmidrule(l){1-2}

        MLP + SPE & estimator = MLP (Normal), n\_estimators = 50 \\ \cmidrule(l){1-2}

        VAE & latent\_dim = 2, batch\_size = 32, learning\_rate = 0.001, KL\_beta = 1.0, encoder\_neuron\_list=[128, 64, 32], decoder\_neuron\_list=[32, 64, 128], activation = ReLU, dropout\_rate = 0.2 \\ \cmidrule(l){1-2}
        
        GAN & epochs = 10, latent\_dim = 121, batch\_size = 32, d\_optimizer = Adam with learning rate = 0.0001 \& beta\_1 = 0.5, g\_optimizer = Adam with learning rate = 0.0001 \& beta\_1 = 0.5  \& clipvalue = 1.0 \\ \cmidrule{1-2}
        
        MO-GAAL & contamination = Proportion of fraud cases in training partition, n\_sub\_generators = 3, learning rate discriminator = 0.01, learning rate generator = 0.0001, epochs = 2 \\ \bottomrule
        \end{tabular}
    \label{tab:hyperparamsKDD}
\end{table}

 \Cref{tab:hyperparams-cc,tab:hyperparamsKDD} present the nine setups per dataset, which include XGB, MLP, VAE, GAN, MO-GAAL, XGB and MLP with ROS and SPE. For XGB, we tested different setups. First, a vanilla XGB model with standard parameters \cite{xgboostdevelopersXGBoostParameters2022}. Second, a XGB model was fine-tuned using \textit{RandomSearchCV} across parameters such as \textit{n\_estimators}, \textit{max\_leaves}, \textit{learning\_rate}, \textit{gamma}, \textit{max\_depth}, \textit{subsample}, \textit{reg\_alpha}, \textit{reg\_lambda}, \textit{scale\_pos\_weight}, with RandomSearch yielding the most promising results on the KDD dataset. These tuned hyperparameters were then compared to the performance of the vanilla XGB model, XGB + ROS and XGB + SPE, and the best performing model was selected for further evaluation.

The setup for the MLPs was taken from \cite{cholletImbalancedClassificationCredit2019}. Furthermore, hyperparameter tuning based on scikit-learn's \textit{RandomSearchCV} was performed for the KDDCUP99 dataset. VAE was trained in an unsupervised fashion on the negative classes to learn the underlying probability distribution, following \cite{bankAutoencoders2021}. GAN was implemented following  \cite{sayakConditionalGAN2021}. GAN and VAE were not combined to ROS or SPE, as these were trained using the information of the negative class only. MO-GAAL was trained using the parameters from \cite{hajekFraudDetectionMobile2023}, whereby \textit{n\_sub\_generators} was reduced to 3 and the number of epochs to 2, as already \cite{hajekFraudDetectionMobile2023} reported large execution time. The python library \textit{PyOD} was used to implement VAE and MO-GAAL \cite{winstonliSourceCodePyod2022}.

%% file: sections/3_results.tex
\section{Results}

\begin{figure}[!htb]
    \centering
    \includegraphics[width=\textwidth]{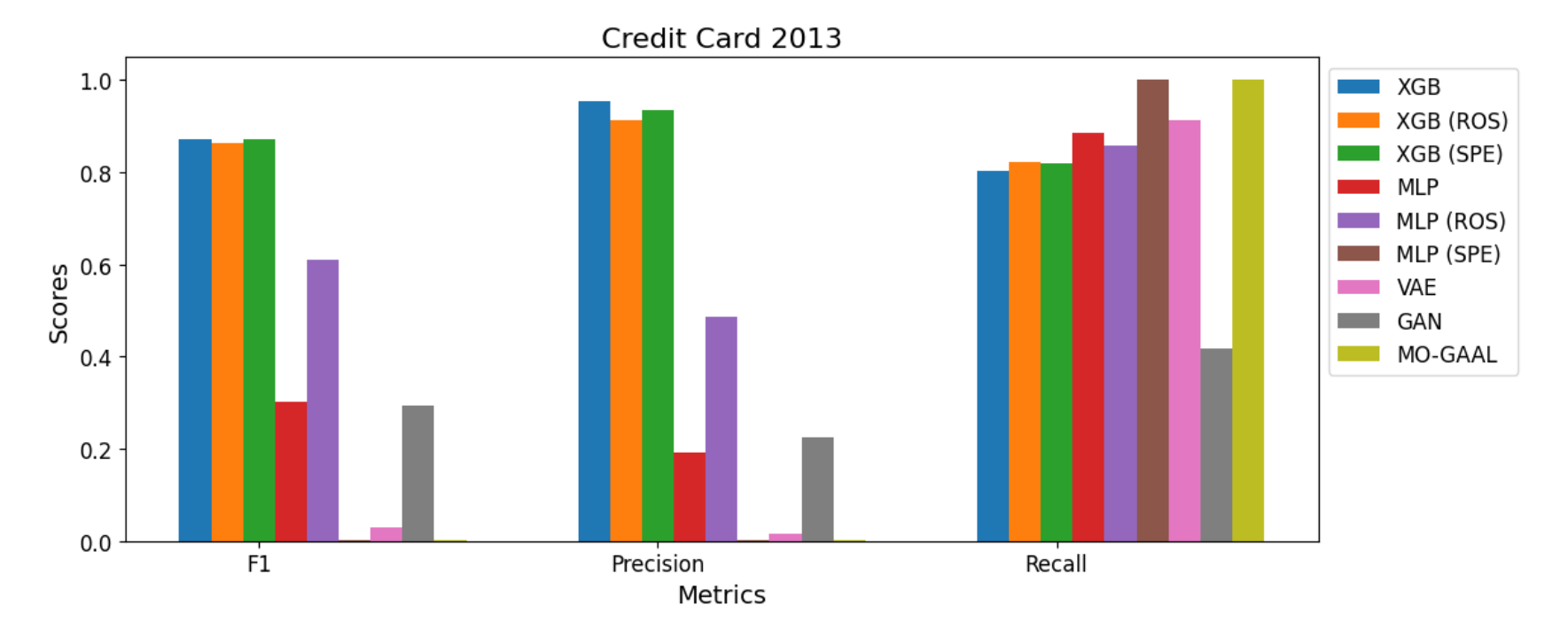}
    \caption[Models' performance on the credit card dataset]{Models' performance on the credit card dataset, based on 5-fold cross-validation on the training data. For XGB and MLP, only the best sampling combinations are shown.}
    \label{fig:model-results-CC}
\end{figure}

\begin{figure}[!htb]
    \centering
    \includegraphics[width=\textwidth]{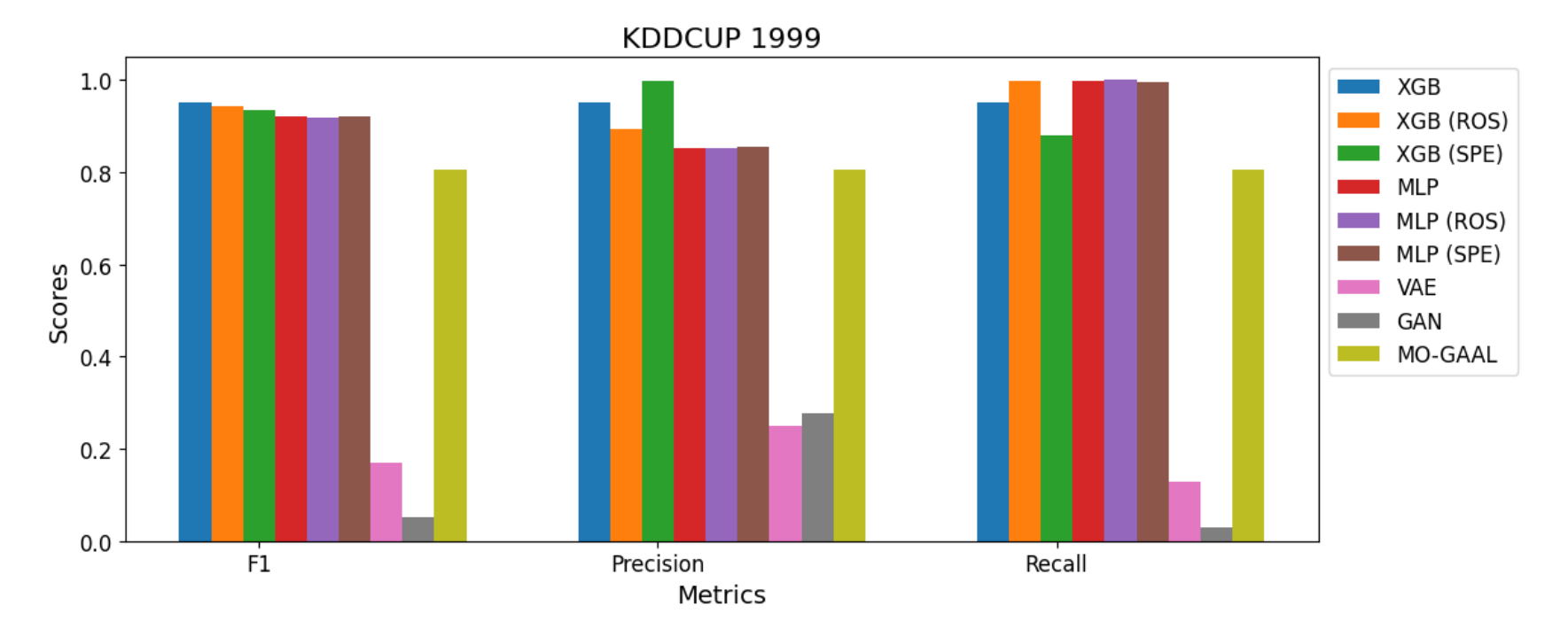}
    \caption[Models' performance on the KDDCUP99 dataset]{Models' performance on the KDDCUP99 dataset, based on 5-fold cross-validation on the training data. For XGB and MLP, only the best sampling combinations are shown.}
    \label{fig:model-results-KDD}
\end{figure}

\begin{figure}[!htb]
    \centering
    \includegraphics[width=\textwidth]{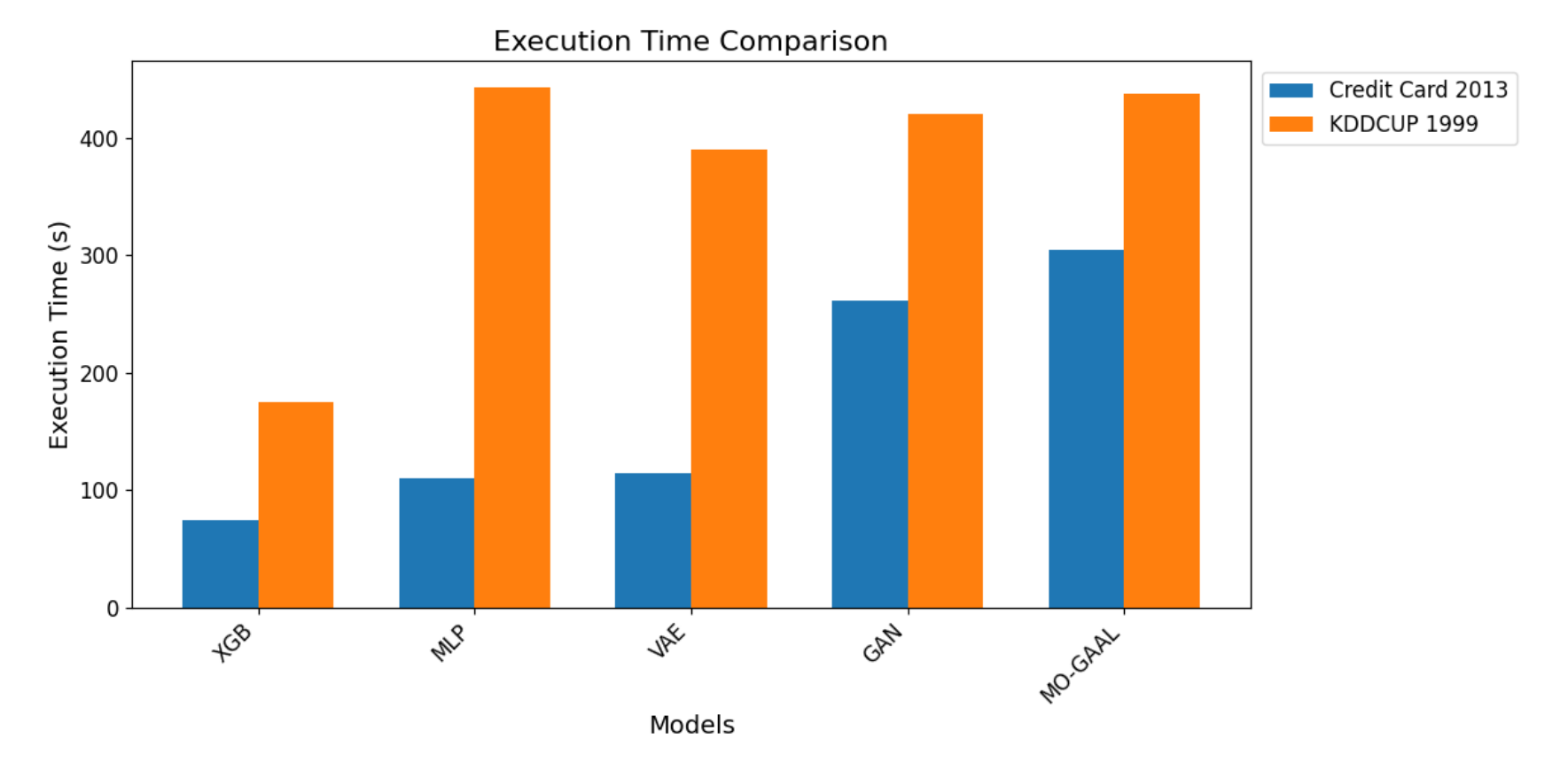}
    \caption[Training times in seconds (Execution Time)]{Training times in seconds (Execution Time). XGB proves to be the most efficient model, while MO-GAAL stands out due to its very long training time. For XGB and MLP, only the best sampling combinations are shown.}
    \label{fig:model-times}
\end{figure}

The models' performance can be seen in \Cref{fig:model-results-CC,fig:model-results-KDD} and their training times in \Cref{fig:model-times}. \Cref{fig:model-results-CC,fig:model-results-KDD} show that XGB outperforms all models and its performance is similar on both datasets. MLP is the second best. MO-GAAL is able to learn and detect anomalies on KDDCUP but not on Credit Card Dataset. VAE and GAN show poorer generalization performance for both datasets. As seen in \Cref{fig:model-times}, it is clear that training on KDDCUP takes longer time than that taken on Credit Card for all models, being XGB the most efficient and MO-GAAL the least.

Furthermore, we observe that the combination of MLP + ROS led to an improvement in anomaly detection for both datasets. Particularly in the case of the credit card dataset, large improvements were achieved by oversampling the positive class.

\begin{figure}[ht!]
    \center
    \includegraphics[width=\textwidth]{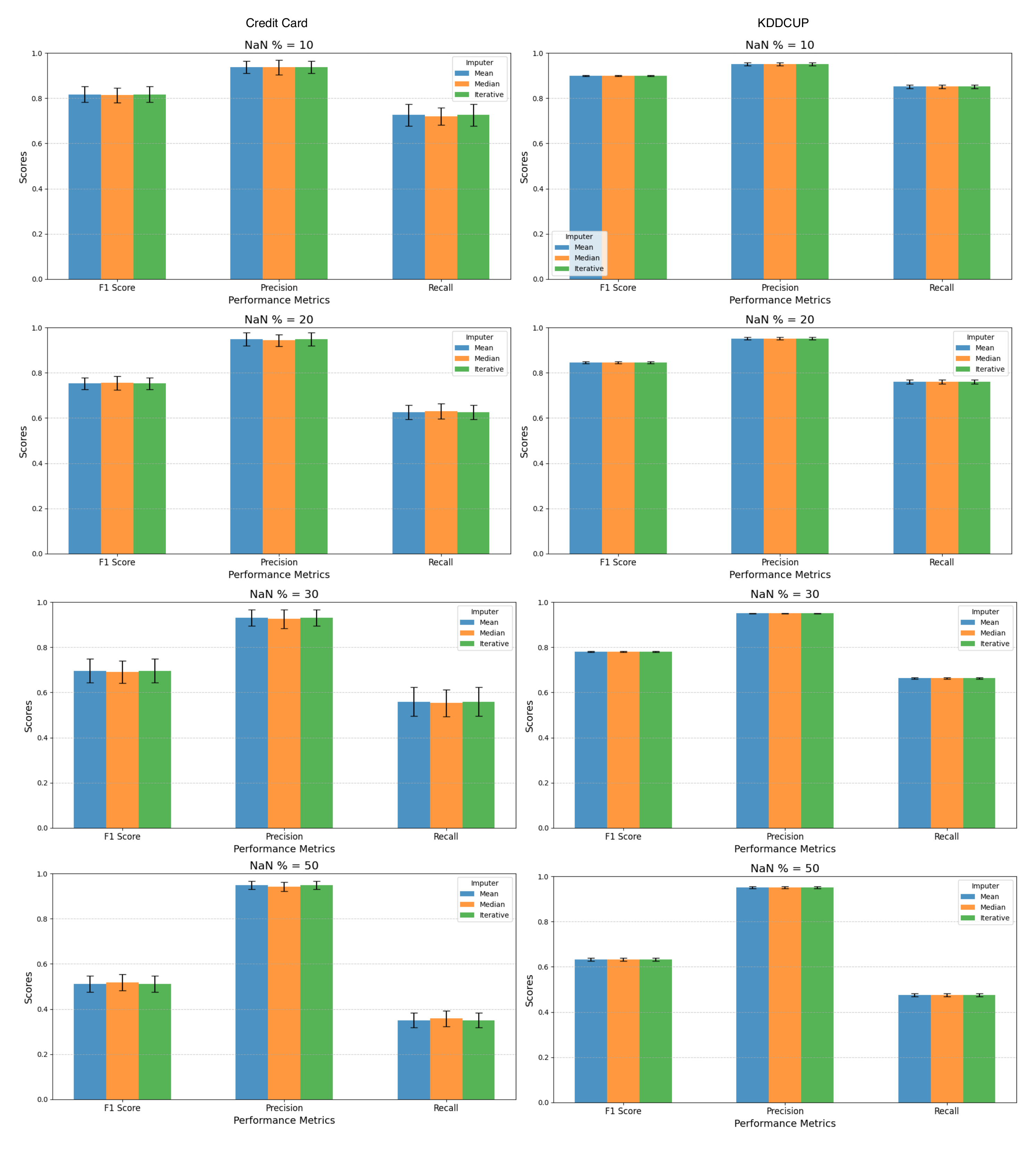}
    \caption{XGB performance with missing data using imputation techniques on credit card and KDDCUP dataset. The error bars represent the standard deviations of the measurements, each of which was performed ten times.}
    \label{fig:impute-times}
  \end{figure}

Finally, since XGB performed best for both datasets, it was used to study the effect of missing data with random deletion of 10, 20, 30, and 50 \% of the data, see results in \Cref{fig:impute-times}.  For the Credit Card dataset, XGB + SPE and for the KDDCUP99 dataset, XGB without any sampling methods were the baseline models (see \Cref{tab:hyperparams-cc,tab:hyperparamsKDD}). Three imputation methods (Mean, Median, IterativeImputer) were evaluated in ten independent runs to avoid coincidences. No significant differences between the imputation methods on both datasets ($p > 0.05, n = 5$) were found.

We also found enormously high execution times for the \textit{IterativeImputer} for the KDDCUP99 dataset.

%% file: sections/4_discussion.tex
\section{Discussion}

This section discusses the main findings.

\subsection{On the general performance}

In preliminary experiments we observed other metrics such as Matthews Correlation Coefficient (MCC)  and Geometric Mean, and noticed that F1, Precision, and Recall are the most relevant to drive conclusions. The results show that XGB and MLP combined to sampling or ensembling achieved the best performance for both datasets. The results confirm the great popularity of XGB in the machine learning community for tabular data. In recent years, the use of XGB has led to many wins in machine learning competitions, with MLPs just behind XGB in terms of popularity \cite[1]{chenXGBoostScalableTree2016}.

ROS and MLP on the credit card dataset produced a jump in \textit{precision}, but this was at the expense of deteriorating \textit{recall}. Exactly the opposite could be observed with Random-Under-Sampling (RUS) in \cite{hajekFraudDetectionMobile2023} in combination with XGB and other models. SPE applied to MLP had a negative effect in performance on the credit card dataset. \cite{liuSelfpacedEnsembleHighly2020} evaluated the performance of MLPs in combination with several sampling methods, including RUS.

We therefore recommended to evaluate several techniques for coping with imbalanced data for each dataset. However, it must also be mentioned that due to the limited time frame of this research, no in-depth experiments with different hyperparameters for SPE were conducted. Besides, future work could investigate classifier performance on different imbalance ratios as it was done in \cite{VELARDE2024200354}.

\subsection{On generative models} \label{pos:discussion-gen-model}

The generative models (VAE, GAN, MO-GAAL) achieved significantly poorer results compared to XGB and MLP. This could be due to their complexity. In particular, the optimization of the hyperparameters for generative models proved to be more difficult, as several MLPs with different objectives were often trained here and the implementation and execution of the optimization for XGB and a simple MLP were significantly simpler. 
Our results are similar to those obtained by \cite{hajekFraudDetectionMobile2023} on another dataset called \textit{PaySim} dataset \cite{Lopez-Rojas2016}.

\subsection{On imputation methods}
  The investigation of the three imputation methods showed that for the \textit{Credit-Card} dataset, the \textit{IterativeImputer} outperforms \textit{Mean} and \textit{Median} imputation methods on 10 \% and 20 \% of missing data. On the KDDCUP99 dataset, there is no difference on whichever method is used for imputation, except for the execution time. \textit{IterativeImputer} posses an enormous processing time, which is particularly evident in the KDDCUP99 dataset. 

The \textit{IterativeImputer} complexity increases with an increasing number of features (columns) and size of the dataset (rows) and impairs the practicability of the method. The complexity $\mathcal{O}$ is defined as $\mathcal{O}(knp^3 \min(n,p))$, where $k$ is the number of maximum iterations, $n$ is the number of entries in the dataset and $p$ is the number of features. It is important to note at this point that the \textit{IterativeImputer} can be adjusted by parameters to counteract a high runtime \cite{scikit-learn}. In this case, however, these parameters were not optimized as this was not possible due to time constraints for this project.

%% file: sections/5_conclusion.tex
\section{Conclusion}
This work contributes to a better understanding of the problem of imbalance classification, considering a systematic evaluation of supervised and unsupervised learning models. We demonstrate on two representative datasets for cybersecurity that supervised learning algorithms outperform unsupervised learning approaches. Moreover, depending on the model and the dataset, some sampling and ensembling techniques can help improve recognition. But these should be used carefully depending on the performance. We observed that Random Oversampling and Self-Paced Ensembling should be tested on each dataset and model before its application. Interestingly, there was no difference on the imputation methods tested (IterativeImputer, mean and median), except that IterativeImputer is not recommended for large datasets due to increased complexity and execution time.

%% file: sections/6_contributions.tex
\section{Author Contributions} M. H. software implementation, experimental design, data analysis, initial paper draft including figures and tables, paper writing and revisions. G.V. supervision, experimental design, paper writing and revisions.